\documentclass[fleqn,10pt]{wlpeerj}
\usepackage{subcaption}
\usepackage{float}
\usepackage{multirow}

\title{Data Safety: Synthetic Data Quality Analysis Using CIFAKE Dataset}

\author[1]{Kuniko Paxton}
\author[2]{Amila Akagić}
\author[1]{Koorosh Aslansefat}
\author[1]{Dhavalkumar Thakker}
\author[1]{Yiannis Papadopoulos}
\affil[1]{School of Digital and Physical Science, University of Hull, Cottingham Road, Hull, HU6 7RX, United Kingdom}
\affil[2]{Faculty of Electrical Engineering, University of Sarajevo, Zmaja od Bosne bb, Sarajevo, 71000, Bosnia and Herzegovina}
\corrauthor[1]{Kuniko Paxton}{k.paxton@hull.ac.uk}

\begin{abstract}
Recently, the societal implementation of high-performance image classification models has expanded rapidly. While these models require vast amounts of training data to improve performance, securing sufficient real images is often impractical. As a means to compensate for this shortage, the use of synthetic data is becoming widespread. However, synthetic images are not necessarily equivalent to real images for training purposes. This study systematically analyzes the differences between two types of synthetic images created by different generation methods and real images from three perspectives: high-dimensional feature space, low-level statistics in color space, and the model training process. Furthermore, it experimentally verifies how synthetic data should be utilized by considering realistic data mixing scenarios.
This enables the proposal of an evaluation and application strategy for performing preliminary assessments on synthetic images of unknown quality and safely incorporating them into training. This research aims to contribute to enhancing the reliability and safety of image classification models utilizing synthetic images.
\end{abstract}

\begin{document}

\flushbottom
\maketitle
\thispagestyle{empty}

\section*{Introduction}
Advancements in computer vision have expanded its applications across diverse domains such as healthcare, manufacturing, surveillance, autonomous driving, and transportation, significantly increasing its societal impact. Given these circumstances, sophisticated deep learning models centered on image classification generally require large amounts of training data. However, in real-world scenarios, constraints such as cost, data privacy, ethical concerns, availability, and annotation burdens make it difficult to collect sufficient images to train models \cite{marwala2023use,jordon2022synthetic,dahmen2019digital}. Therefore, in recent years, efforts to utilize generated data for training have become widespread as a means to compensate for the shortage of real image data. Generation technologies based on GANs \cite{goodfellow2020generative}, diffusion models \cite{ho2020denoising}, and variational autoencoders \cite{kingma2013auto} have enabled the creation of high-quality images that are difficult for human perception to distinguish from real ones.

However, the use of synthetic data is not necessarily safe. For example, research by \cite{akagic2024exploring} reported that while models trained on synthetic images showed comparable accuracy to those trained on real images, significant differences emerged in metrics reflecting class imbalance and misclassification tendencies, such as F1 scores by validating the CIFAKE dataset \cite{bird2024cifake,krizhevsky2009learning}. CIFAKE contains 60,000 synthetic images generated by Stable-Diffusion-v1-4 \cite{rombach2022high} and 60,000 real images, which were collected from the CIFAR-10 dataset \cite{krizhevsky2009learning} with ten classes. This finding suggests that, contrary to the goal of improving performance through synthetic data, classification performance may degrade when real-world images are input after model deployment, potentially compromising model reliability and safety. A survey for the overall picture of synthetic data research also concluded that synthetic data is convenient, but there are some risks if its quality is not checked \cite{hao2024synthetic}. Therefore, to utilize synthetic data in a manner suitable for real-world deployment, it is crucial to quantitatively understand how generative data differs from real data. This requires a multifaceted analysis of how these differences impact the training process and the correlation of image quality differences and performance.

Previous research on the differences between real and synthetic data using CIFAKE has primarily been developed with the goal of deepfake detection \cite{weir2024enhancing,jagam2025deepfake,das2025edge,mouna2024ai,akram2025comparative,singh2024deep,nourji2025evaluating,jiang2025addressing}. These studies typically formulated the problem as a binary classification task (real or fake) and employed approaches where ML models are trained to recognize patterns that distinguish between the two. However, since the primary focus was on improving detection accuracy, there was often insufficient in-depth analysis of what specifically caused these differences in datasets. While this is crucial for addressing societal issues arising from misuse, its objectives differ from those of evaluation and analysis frameworks designed to safely use of synthetic data as a trainable resource. The objective of this research is not to detect generated images, but to establish a framework for safely using generative images through multifaceted evaluation methods. To that end, we analyze differences between generated and real data based on information within the generated data, including cases where the generation model's quality is unknown and is not systematically accounted for in model training.

Specifically, using three heterogeneous characteristics datasets,  the CIFAR-10, CIFAKE dataset alongside images generated by fine-tuning with Elucidating the Design Space of Diffusion-Based Generative Models (EDM) \cite{karras2022elucidating} which is a state-of-the-art model, we analyze three main perspectives: (1) High-dimensional feature space, (2) low-level statistics measures, and (3) model structures in the learning process. This study proposes a prior validation framework for safely utilizing synthetic data in training, based on multifaceted analysis results. This framework identifies metrics that quantify the distribution gap between generative and real data and estimates the required proportion of real data to blend to achieve the target performance, based on the metrics. This clarifies the previously unquantified acceptable usage range for synthetic data, contributing to improved AI safety in real-world operations.

\section*{Related Work}
This section explores research focusing on the differences between real and synthetic images. Prior research can be categorized into four main areas.

\subsection*{Model-centric analysis}
These studies focus on how differences between real and synthetic data affect model behavior. For example, in a distillation learning setting, \cite{hennicke2024mind} analyzed which layers of student models trained on synthetic data were most affected, clarifying factors behind performance gaps compared to real data. Diffusion curriculum \cite{liang2025diffusion} was designed using image guidance strength, improving synthetic data learning efficiency by controlling the generation process. While these studies provide valuable insights into model behavior, systematically analyzing the distribution differences in the data itself is limited. Therefore, this manuscript focuses on a multifaceted analysis of the data itself and examines how it correlates with the model's learning process.

\subsection*{Learning process strategy}
The following group of studies verifies the effectiveness of learning methods using synthetic data. \cite{wachter2025development} systematically compared multiple mixing strategies from the perspective of how synthetic and real data should be mixed. Through empirical outcomes, they presented practical implications for mixed learning. However, they did not provide quantitative operational metrics, such as how much real data must be mixed to enter a safe zone. \cite{singh2024synthetic} conducted multifaceted robustness evaluations on unsupervised, self-supervised, multi-modal models trained solely on synthetic data, examining aspects such as shape bias, background bias, and noise resilience. \cite{he2022synthetic} experimentally verified the effectiveness of synthetic data in zero-shot, few-shot settings, and pre-training. They also delved into the question of which parts of the model structure are influenced by learning with synthetic data. In \cite{yuan2023real}, performance was evaluated across three settings: synthetic only, real + synthetic augmentation, and synthetic scaling law. However, both studies by \cite{he2022synthetic} and \cite{yuan2023real} provided limited analysis of data quality at the data level and of its direct links to performance degradation factors.

\subsection*{Data-centric analysis}
An analysis by \cite{yuan2023real} centered on distribution consistency in high-dimensional representation spaces based on the feature space of the CLIP image encoder. However, they did not directly analyze pixel-level statistical properties or differences in low-level visual elements. \cite{geng2024unmet} highlights that generative models can output unnatural quirks or artificial noise imperceptible to humans, and inaccurately represent important visual details. They demonstrate that these issues can stem from models learning unnecessary features or failing to sufficiently learn essential features. However, the specific data distribution discrepancies causing these differences are not always systematically analyzed. \cite{babbar2024different} aimed to explain differences between datasets. They summarized differences in essential features across tasks using feature importance vectors and defined their own metrics. However, they did not delve into quantitative evaluations of effect sizes, such as how much individual feature misalignments impact classification performance. pyMDMA \cite{faccoco2025pymdma} is a tool that systematizes quality evaluation metrics for real and synthetic data and implements them as an audit framework. However, verification of how these quality metrics actually impact the learning process and performance remains limited at this stage.

\subsection*{Data Analysis using CIFAKE for Deepfake Detection}
The following three studies all primarily aim to detect deepfakes, but they also analyze data characteristics. The studies by \cite{mehta2025enhancing} and \cite{mehta2024swin} analyzed distributions in RGB and YCrCb using Kullback-Leibler divergence, introducing a preprocessing step that utilizes these differences for feature extraction. It focuses on the unnatural color often seen in generated images, skewed color difference distributions in the relationship between luminance and color components. Chrominance statistics can effectively be used for authenticity discrimination tasks, such as real vs. fake. However, in semantic classification tasks, deviations in hue component statistics are not necessarily intrinsic features. Additionally, \cite{hosencifake} applied LIME to deepfake detection models, analyzing them through the explainability of prediction results. However, this primarily visualizes the model's reasoning basis and differs in nature from approaches that systematically and quantitatively analyze the image data distribution itself.

Overall, while prior studies have explored differences between real and synthetic data from model-centric, learning process strategy, and data-centric perspectives, a comprehensive framework that systematically connects multi-level distribution discrepancies, ranging from high-dimensional feature spaces to low-level statistics, with downstream classification performance remains underdeveloped. This gap motivates this study.

\subsection*{Research Questions} \label{sec:research_questions}
Given the above introduction and gaps in related research, this study sets the following research questions to clarify the differences between generated data with distinct characteristics. \textbf{CIFAKE1} denotes generated data that fails to train models capable of correctly classifying real images, whereas \textbf{CIFAKE2} denotes generated data that successfully enables such models.
\\
\begin{itemize}
    \item \textbf{RQ1:} What differences arise in the high-dimensional feature space pre-trained on general images between CIFAKE1 and CIFAKE2?
    \item \textbf{RQ2:} What differences can be observed in the low-level statistical feature space between CIFAKE1 and CIFAKE2?
    \item \textbf{RQ3:} What differences emerge in the learning process between CIFAKE1 and CIFAKE2?
    \item \textbf{RQ4:} How can CIFAKE1 be effectively utilized in the learning process?
\end{itemize}

\subsection*{Contributions}
By responding to the research questions above, this study comprehensively analyzes differences in characteristics between the two types of synthetic datasets: CIFAKE1 and CIFAKE2. As a result, it highlights the potential risks of generative data that have not been sufficiently discussed to date, while also providing guidelines for its safe utilization. Moreover, by clarifying appropriate usage methods based on the characteristics of generative data, it contributes to enhancing the safety of learning using synthetic data.

\section*{Methodology and Experimental Settings}
\begin{figure*}
    \centering
    \includegraphics[width=1\linewidth]{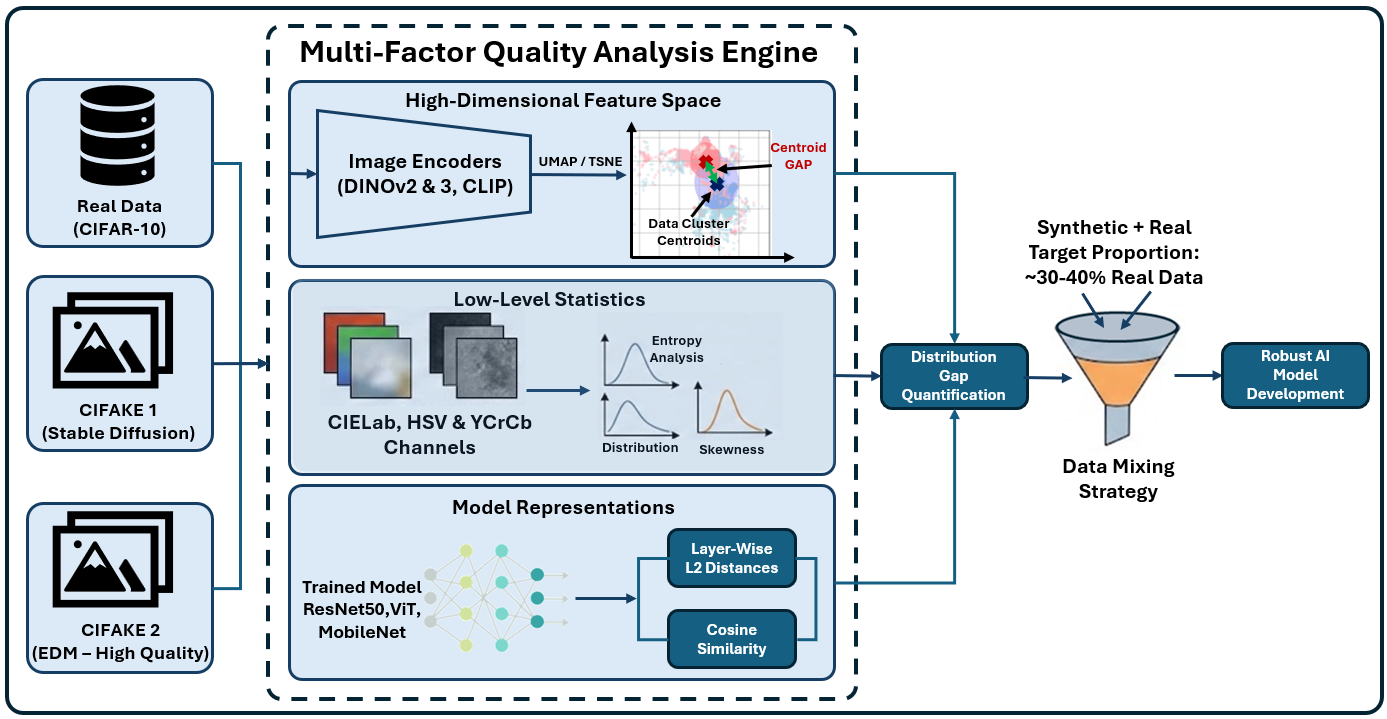}
    \caption{Research Overview and Proposed Synthetic Data Analysis Framework: Evaluation of three heterogeneous datasets using the proposed multi-factor quality analysis engine, and assessment of the impact of the resulting synthetic data in realistic scenarios.}
    \label{fig:overview}
\end{figure*}

Figure \ref{fig:overview} presents the data quality analysis pipeline, where three heterogeneous datasets are analyzed using our proposed framework, and the impact of the synthesized data is examined under realistic conditions. The details are explained in each subsection.

\subsection*{Datasets}
The datasets used in this study are the following three types in Table \ref{tab:dataset_information}. 10,000 records from the REAL dataset are used as test data, while CIFAKE1 and CIFAKE2 are used solely for training. The number of images per class is set equally across all 10 classes.

\begin{table}[H]
\resizebox{\columnwidth}{!}{%
\begin{tabular}{l|l| p{0.75\columnwidth}}
\hline
Data Name & Num & Description \\ \hline
REAL & 60,000 & CIFAR-10 \\ \hline
CIFAKE1 & 60,000 & CIFAKE10 generated by Stable-Diffusion-v1-4 \\ \hline
CIFAKE2 & 60,000 & Images generatied by EDM  \\ \hline
\end{tabular}}
\caption{Dataset Information}
\label{tab:dataset_information}
\end{table}

\subsection*{Classification Models}
Three types were adopted: the CNN-based ResNet50 \cite{he2016deep}, the MobileNetV2 \cite{sandler2018mobilenetv2} designed for mobile environments, and the transformer-based Vision Transformer \cite{dosovitskiy2020image}. All models were pre-trained on ImageNet \cite{deng2009imagenet}, and this study focuses on fine-tuning these models. This is because, in many image classification applications, it is common practice to utilize pre-trained backbones rather than training models from scratch. Training was conducted for up to 50 epochs, with the initial learning rate set to 0.1 and gradually reduced to achieve convergence. For each experiment, the model yielding the highest F1 score is selected for analysis. Data augmentation is limited to shuffling, as image color-space analysis is conducted.

\subsection*{Differences in High-Dimensional Feature Space}
To evaluate differences in image representations within high-dimensional feature spaces, this study employs three feature extraction models: Contrastive Language-Image Pre-training (CLIP) \cite{radford2021learning}, DETRwith Improved deNoising anchOr
boxes v2 (DINOv2) \cite{oquab2023dinov2}, and DINOv3 \cite{simeoni2025dinov3}. To visualize differences in the distribution of feature representations obtained by these models, plots are generated using Uniform Manifold Approximation and Projection (UMAP) \cite{mcinnes2018umap} and t-Distributed Stochastic Neighbor Embedding (t-SNE) \cite{maaten2008visualizing} as dimensionality-reduction techniques. Total six patterns are analyzed.

\subsection*{Differences in Low-Level Statistics}
In analyses of statistical differences in low-level features, illuminance components, the V channel in the HSV color space, the Y channel in the YCrCb color space, and the L* channel in the CIELab color space are targeted. While color components are expected to differ between classes, illuminance components are class-independent and are ideally assumed to be constant. From the first to fourth order statistics and entropy are measured to compare the REAL dataset.

\subsection*{Differences in Model Learning Process}
This section analyzes the following two points: (1) Compares differences in internal representations within the model during the learning process between REAL, CIFAKE1, and CIFAKE2 based on cosine similarity and L2 difference per layer. (2) Systematically evaluates the effects of mixing generated images based on realistic usage scenarios.
\begin{enumerate}
    \item \textbf{One class replacement:} For instance, in cases where data for specific disease labels is extremely scarce, such as skin lesion images, methods that supplement only the minority class with generative images are commonly employed. Therefore, this experiment replaced only the class showing the most significant performance degradation with REAL images to analyze its impact.
    \item \textbf{Randomized mixture: } This setting anticipates situations where sufficient real images are unable to be collected due to rare events, and generates images mixed across all classes. In this verification, the mixing ratio of REAL images is gradually varied evenly from 10\% to 90\% across each class to analyze the impact of the mixing ratio on model performance.
    \item \textbf{Additional finetuning with small real data:} This configuration assumes a scenario where a small amount of data is available locally and models have already been trained on generated images. It investigates the differences between models trained by fine-tuning the best F1 score model trained on CIFAKE1 with an additional 10\% of REAL images, for periods ranging from 10 epochs up to a maximum of 100 epochs.
\end{enumerate}

\section*{Results}
\subsection*{Generative Image vs Real Image}
\begin{figure*}
    \centering
    \includegraphics[width=1\linewidth]{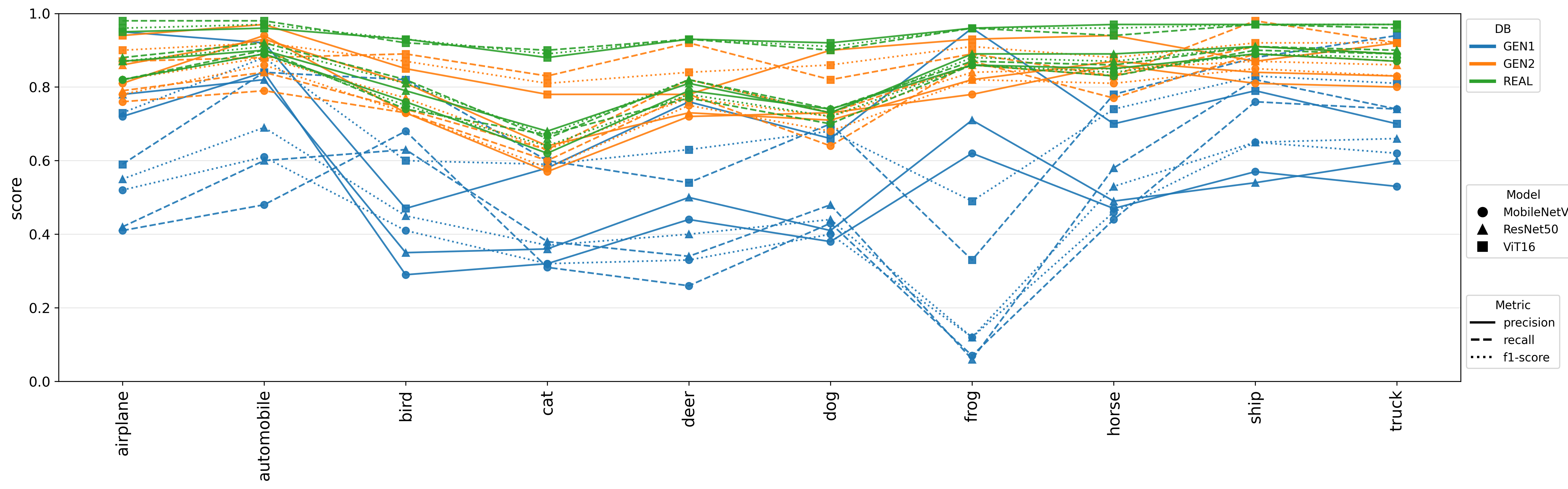}
    \caption{Performance of Each Class and Model: This figure compares performance across classes for the three datasets: CIFAKE1, CIFAKE2, and REAL. Datasets are indicated by color, models by markers, and performance metrics (Precision, Recall, F1-score) by line type.}
    \label{fig:performance_each_class}
\end{figure*}

As shown in Figure \ref{fig:performance_each_class}, CIFAKE1 performs worse than CIFAKE2 and REAL across many classes and models across all evaluation metrics. This result suggests that the CIFAKE1 dataset possesses different characteristics from the other two. On the other hand, CIFAKE2 shows a trend largely similar to REAL, confirming that it has properties close to REAL.

\begin{figure}[t]
    \centering
    \subfloat[CIFAKE1\label{fig:class_heatmap_FAKE1}]{
        \includegraphics[width=1\linewidth]{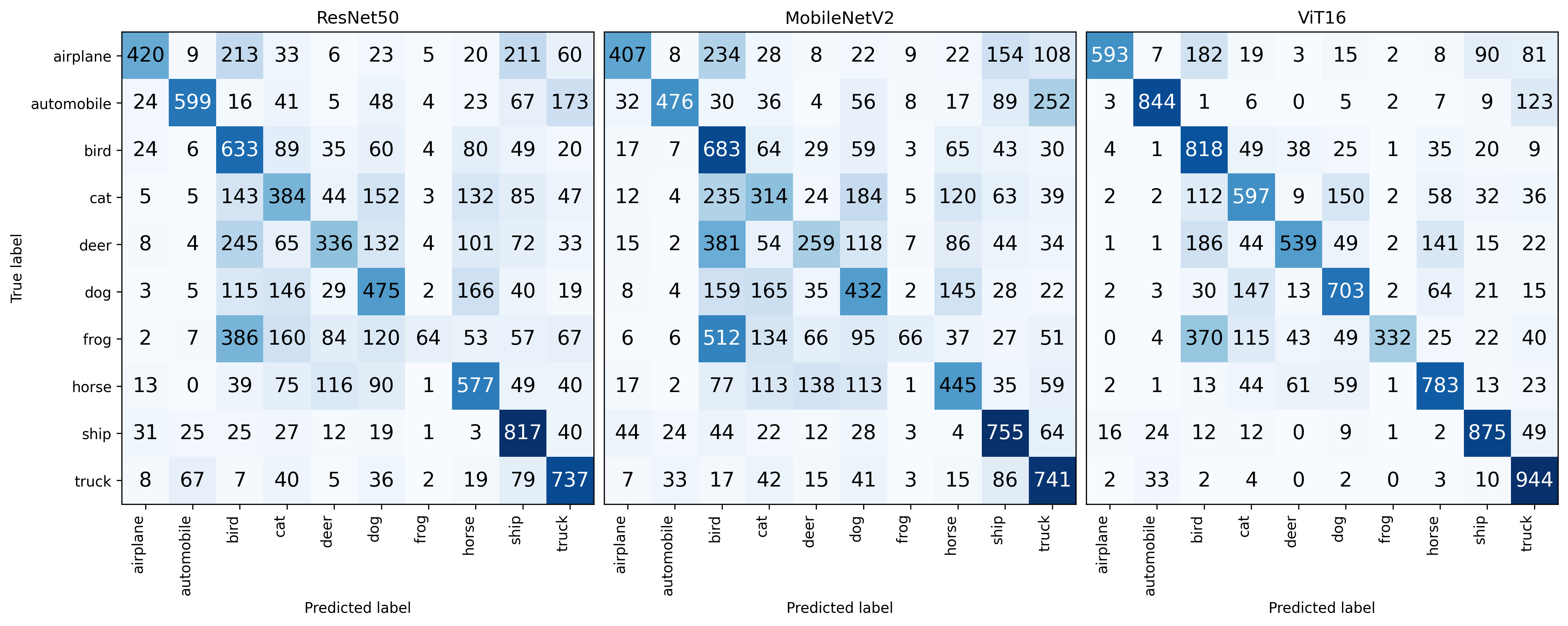}
    }
    \hfill
    \subfloat[CIFAKE2\label{fig:class_heatmap_FAKE2}]{
        \includegraphics[width=1\linewidth]{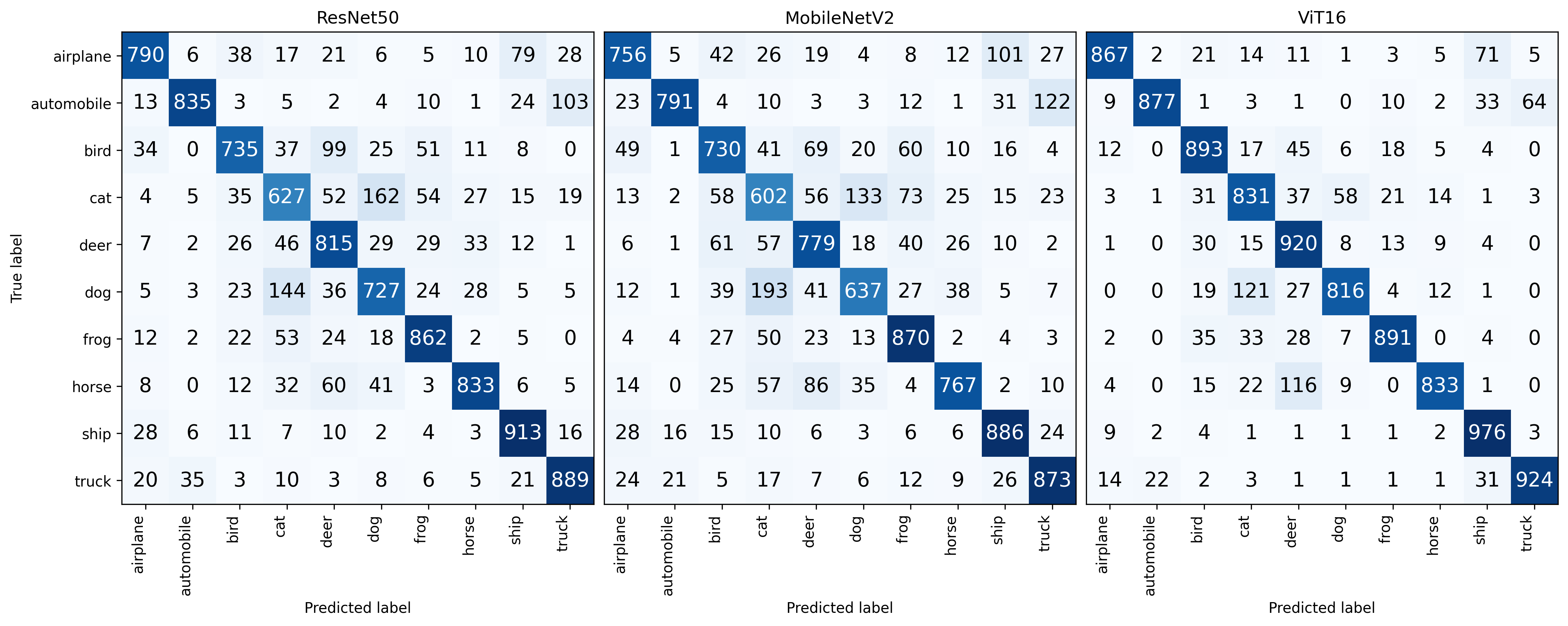}
    }
    \caption{Confusion matrices for classification results by class in the CIFAKE datasets. The Y-axis represents the true class and the X-axis represents the predicted class.}
    \label{fig:class_heatmap}
\end{figure}

Furthermore, examining the classification details of CIFAKE1 in the confusion matrix (Figure \ref{fig:class_heatmap_FAKE1}) reveals that all models tend to misclassify frog as bird and cat. Cats and deer are prone to misclassification as birds, horses, or dogs, while CNN-based models showed a tendency for dogs to be misclassified as birds, cats, or horses. On the other hand, trucks and ships generally exhibited relatively fewer misclassifications overall. As shown in Figure \ref{fig:class_heatmap_FAKE2}, CIFAKE2 showed a similar trend but, due to its higher performance, did not exhibit the same level of disturbance as CIFAKE1.

\subsection*{Differences in High-Dimensional Feature Space}
\begin{figure*}[t]
    \centering
    \subfloat{%
        \includegraphics[width=0.9\linewidth]{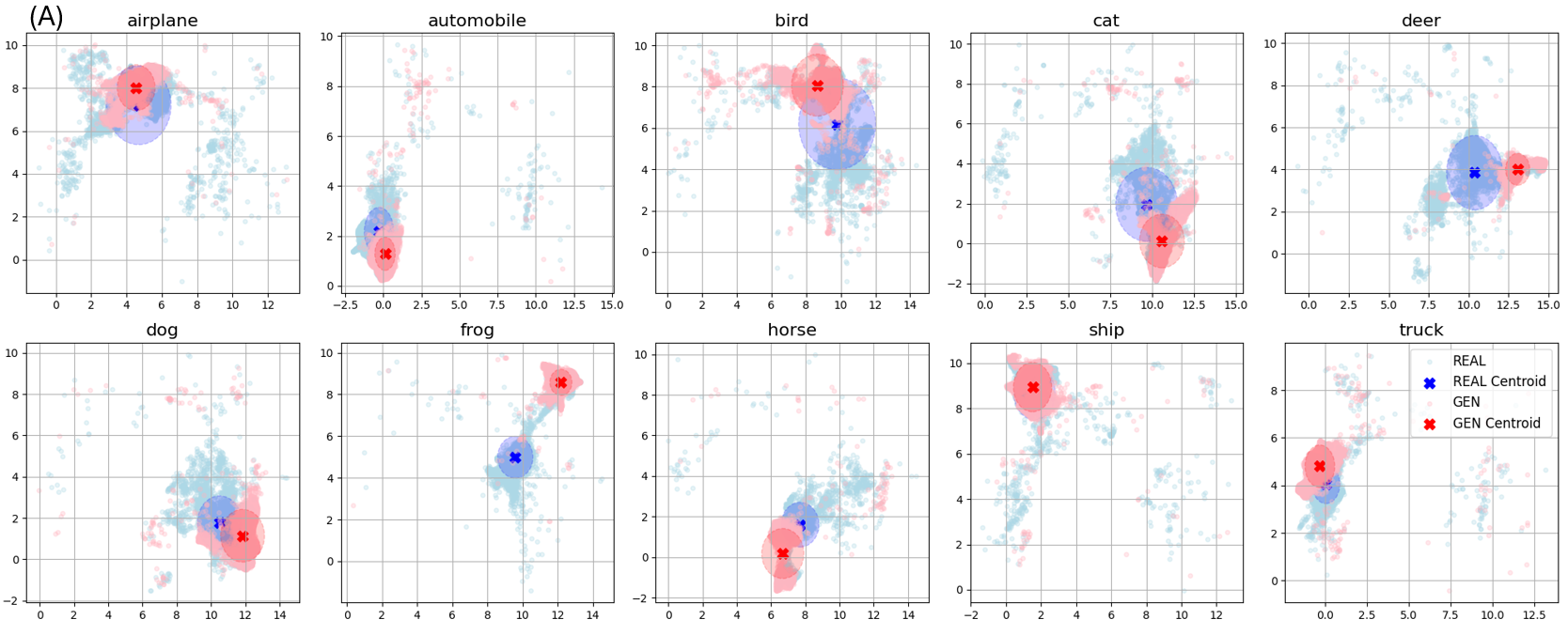}
        \label{fig:umap_CIFAKE1}
    }\par
    \subfloat{%
        \includegraphics[width=0.9\linewidth]{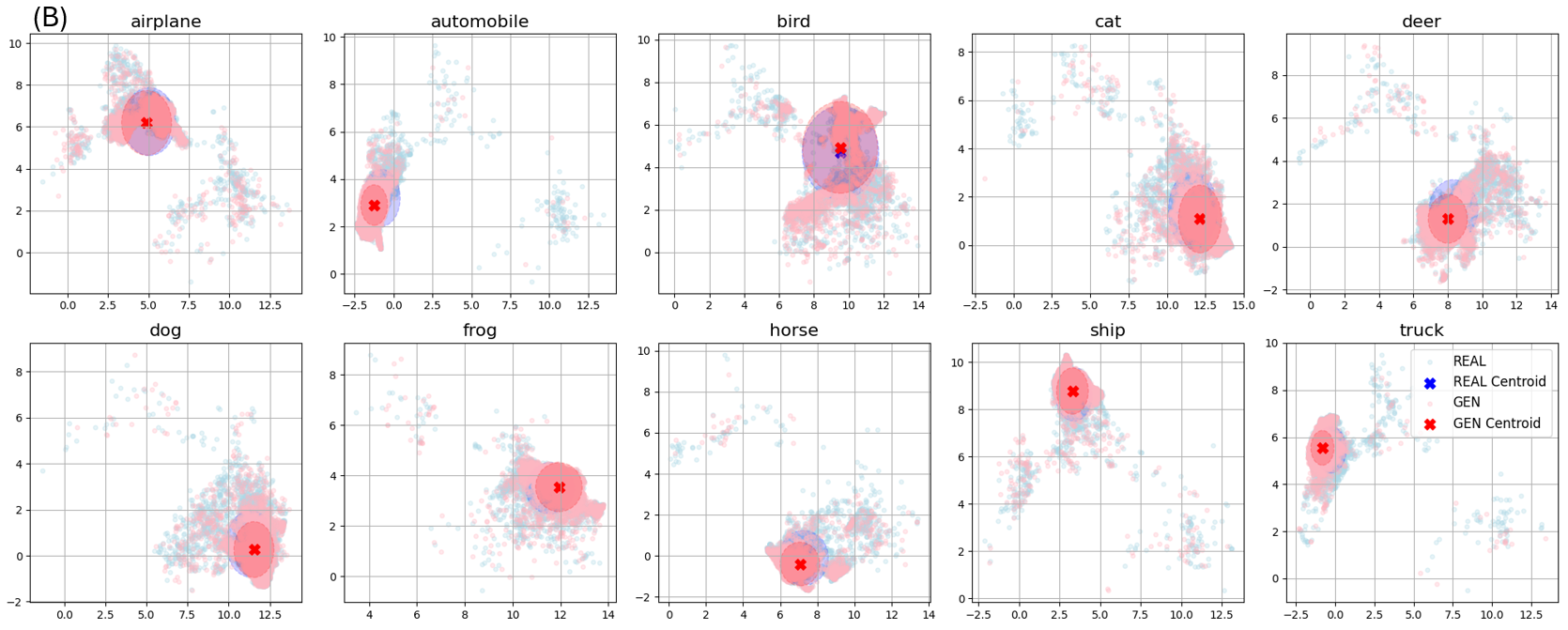}
        \label{fig:umap_CIFAKE2}
    }
    \caption{UMAP visualization results of DINOv3 image embeddings in class-specific generated datasets. (A) is created using CIFAKE1 data, and (B) using CIFAKE2 data. Light-blue data points represent real images, whereas red data points represent generated images. The crosses indicate the centroids of the data points within each class and dataset.}
    \label{fig:umap_dinov3}
\end{figure*}

Visualization results in Figure \ref{fig:umap_dinov3} obtained by projecting the high-dimensional feature space by DINOv3 embeddings onto two dimensions using UMAP reveal that CIFAKE1 tends to cluster relatively far from the centroids of the real data in the UMAP space. This separation is particularly pronounced for frog, bird, and cat, where classification performance degradation was significant, showing large distances between the CIFAKE1 and REAL clusters of centroid distances. In contrast, the centroids of CIFAKE2 are closer to those of REAL, and its data points are generally distributed within the same areas. Furthermore, the size of the circles representing the spread of the distribution is similar, suggesting that the feature distribution in CIFAKE2 is closer to REAL.

In the Table \ref{tab:distance_centroid}, the top table shows results when dimensionality reduction was performed using UMAP, while the bottom row shows results when it was performed using TSNE. Each value represents the distance between the centroids of real and generative data. Notably, large distance differences were observed in the frog class, so the corresponding values are shown in bold.

\begin{table}[]
\begin{tabular}{l|lll|lll|lll|lll}
\hline \hline
\multirow{3}{*}{Label} & \multicolumn{6}{l|}{UMAP} & \multicolumn{6}{l}{TSNE} \\ \cline{2-13}
& \multicolumn{3}{l|}{CIFAKE1} & \multicolumn{3}{l|}{CIFAKE2} & \multicolumn{3}{l|}{CIFAKE1} & \multicolumn{3}{l}{CIFAKE2} \\
 & CL   & V2   & V3  & CL   & V2   & V3  & CL   & V2   & V3  & CL   & V2   & V3  \\ \hline
airplane & 1.39 & 0.81 & 0.84 & 1.19 & 0.18   & 0.15 & 21.68  & 17.49  & 12.12  & 21.47 & 2.92 & 1.72 \\
automobile & 1.39 & 0.76 & 1.23 & 1.82   & 0.31   & 0.56 & 19.98  & 6.02 & 19.66  & 26.42  & 5.67 & 7.36 \\
bird & 1.48   & 2.57   & 2.18   & 1.60   & 0.25   & 0.26 & 23.95  & 25.82  & 31.17  & 29.69  & 5.17   & 3.80  \\
cat &  2.30   & 1.50   & 1.93   & 2.35   & 0.50   & 0.56 & 29.48  & 19.27  & 22.44  & 38.13  & 7.29   & 9.00 \\
deer & 1.67   & 1.67   & 3.69   & 2.29   & 0.53   & 0.52 & 29.10  & 22.90  & 30.19  & 35.03  & 7.40   & 8.56 \\
dog & 1.58   & 1.42   & 1.47   & 1.90   & 0.28   & 0.36 & 21.56  & 22.62  & 19.14  & 28.07  & 3.84   & 5.06 \\
\textbf{frog} & \textbf{5.15} & \textbf{7.47} & \textbf{3.60} & \textbf{1.89} & \textbf{0.41} & \textbf{0.38} & \textbf{57.01}  & \textbf{34.68}  & \textbf{34.84} & \textbf{34.29} & \textbf{6.66} & \textbf{5.89} \\
horse & 1.35   & 1.84   & 1.93   & 2.26   & 0.48   & 0.35 & 21.01  & 22.22  & 25.52  & 31.33  & 6.79   & 4.85 \\
ship & 0.87   & 0.33   & 0.30   & 1.00   & 0.13   & 0.16 & 12.74  & 4.56   & 4.35   & 17.02  & 1.80   & 2.01 \\
truck & 1.21   & 0.67   & 0.89   & 1.82   & 0.44   & 0.52  & 21.26  & 11.01  & 5.67   & 26.14  & 6.60   & 5.71   \\ \hline \hline
\end{tabular}
\caption{Distance from Real Centroid, CL: CLIP, V2: DINOv2, V3: DINOv3}
\label{tab:distance_centroid}
\end{table}

In both DINOv2 and DINOv3, the CIFAKE2 classes consistently approach the real centroids in both UMAP and t-SNE, suggesting that CIFAKE2 is less in overall distribution misalignment. In contrast, CLIP showed no uniform shortening of distances for CIFAKE2, with some classes exhibiting increases and others decreases. That said, frog exhibited significantly larger distances than the other classes in both methods. It was confirmed that classes showing performance degradation also exhibited greater divergence in the feature space.

\begin{figure*}
    \centering
    \includegraphics[width=1\linewidth]{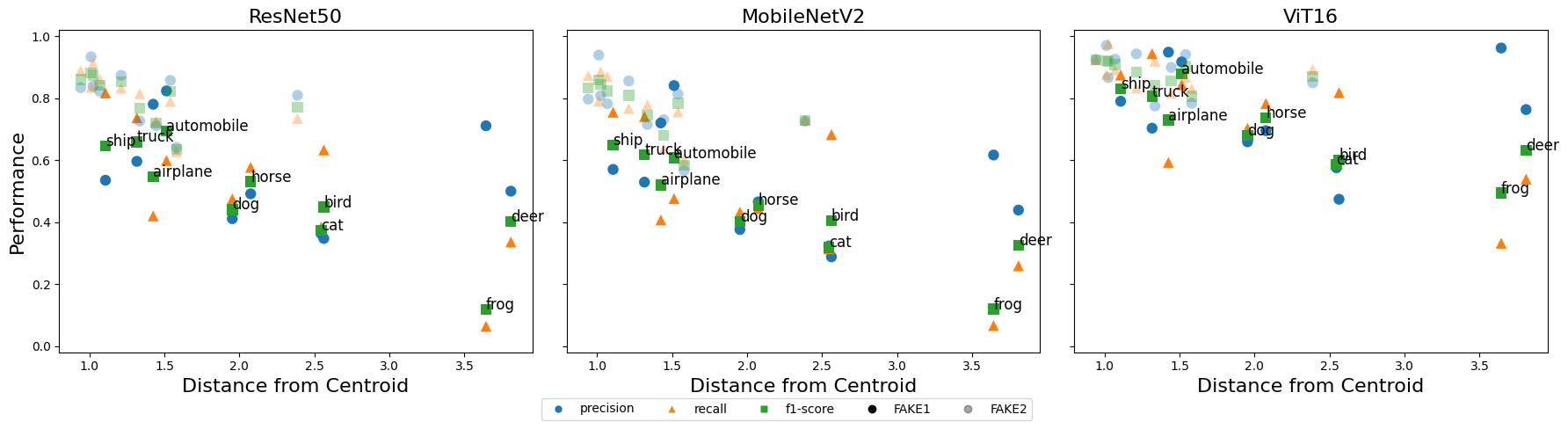}
    \caption{Correlation Between Performance and Distance from Centroid Each Model using DINOv3 and UMAP: Circles, triangles, and squares indicate, respectively, precision, recall, and F1 score. CIFAKE1 represents the dark reference color, while CIFAKE2 represents the light color in the same hue.}
    \label{fig:DINOv3_UMAP_prec_recall_f1}
\end{figure*}

Furthermore, the correlation between distance and performance is shown in Figure \ref{fig:DINOv3_UMAP_prec_recall_f1}. In CIFAKE1, a tendency toward performance degradation was observed as the distance from the real data centroid increased. Meanwhile, even in CIFAKE2, where performance degradation was not pronounced, a tendency toward performance decline with increasing distance was confirmed. However, the impact was relatively minor in CIFAKE2. These results suggest that deviation from the real distribution in the feature space may be associated with performance degradation.

\subsection*{Differences in Low-Level Statistics}
\begin{figure*}
    \centering
    \includegraphics[width=1\linewidth]{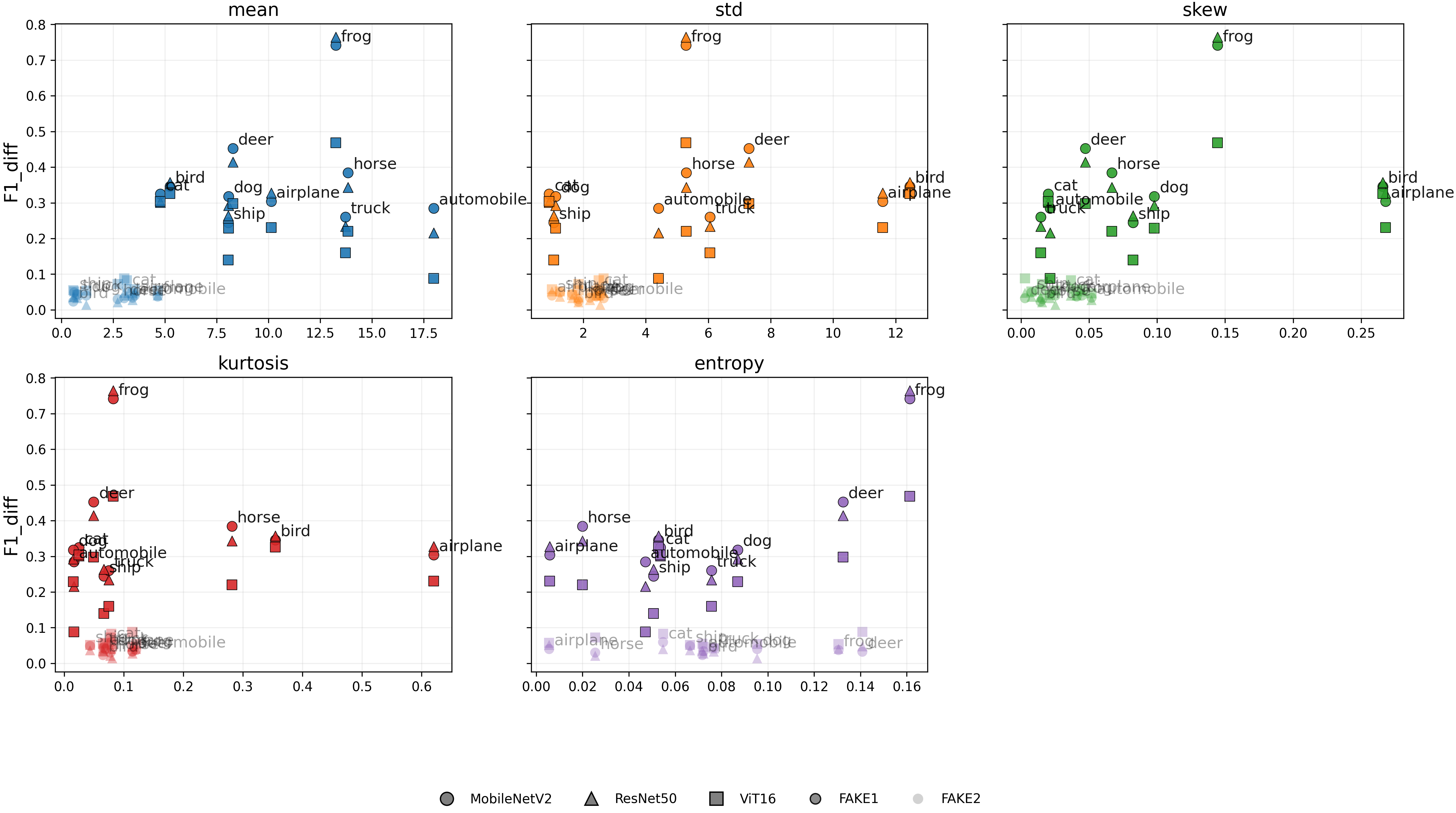}
    \caption{Relationship Between F1 Score and Illuminance (Y-channel) Distribution Differences Each Class: The x-axis represents the magnitude of differences in the statistical distribution, while the y-axis shows the performance degradation relative to training on real data. Marker types indicate different models, and color intensity denotes differences in data generation. Dark colors correspond to CIFAKE1, while light colors correspond to CIFAKE2.}
    \label{fig:gap_Y_F1_diff_FAKE1_FAKE2}
\end{figure*}

As shown in Figure \ref{fig:gap_Y_F1_diff_FAKE1_FAKE2}, analysis of the relationship between distribution differences in real data and performance degradation, focusing on the illuminance channel, revealed that for CIFAKE1, larger distribution differences tended to lead to greater performance degradation across statistical metrics and entropy. However, krutosis was not sufficiently captured. While the linear relationships were relatively weak for the first- and second-order statistics, a clearer linear correlation was observed for skewness and entropy. Furthermore, since CIFAKE2 itself has a small distribution difference from the real data, its impact on performance was also limited.

Furthermore, Table \ref{tab:correlation_illumination} quantitatively evaluated the correlation using Pearson's correlation coefficient, based on the linear trend observed in Figure \ref{fig:gap_Y_F1_diff_FAKE1_FAKE2}. Entropy (indicated in bold) showed a moderate to strong positive correlation with the F1 score and Recall across all models and color spaces when compared to the Illuminance channel. Notably, MobileNet exhibited a strong positive correlation exceeding 0.75 with the F1 score, reaching the highest value of 0.87 with the L channel. While ViT16 showed slightly weaker correlations compared to CNN-based models, it still exhibited positive correlations with the L channel (0.65 for F1 score and 0.75 for Recall) and a negative correlation with Precision (-0.45). These results suggest that changes in the entropy distribution within the L channel significantly impact performance.

\begin{table*}[]
\centering
\resizebox{\textwidth}{!}{%
\begin{tabular}{ll|lllll|lllll|lllll}
\hline \hline
\multirow{2}{*}{Model} & \multirow{2}{*}{} & \multicolumn{5}{l|}{HSV (V)}                     & \multicolumn{5}{l|}{CIELab(L)}                     & \multicolumn{5}{l}{YCrCb(Y)}                      \\
                       &  & Mean  & Std   & \textbf{Ent.} & Kurt. & Skew  & Mean  & Std  & \textbf{Ent.} & Kurt. & Skew  & Mean  & Std   & \textbf{Ent.} & Kurt. & Skew  \\ \hline
MobileNet & F1 & -0.17 & 0.13   & \textbf{0.76} & 0.02 & 0.10 & -0.05 & 0.04 & \textbf{0.87} & -0.16 & 0.17  & 0.13  & 0.10  & \textbf{0.75} & -0.09 & 0.15  \\
MobileNet & Pre & -0.37 & 0.23  & \textbf{0.04} & -0.03 & -0.60 & -0.53 & 0.14 & \textbf{-0.01} & -0.09 & -0.09 & -0.57 & 0.01 & \textbf{0.18} & -0.11 & 0.05  \\
MobileNet & Re  & 0.06  & -0.08 & \textbf{0.68} & -0.05 & 0.36 & 0.22  & -0.13 & \textbf{0.74} & -0.16 & 0.02 & 0.38 & -0.05 & \textbf{0.54} & -0.10 & -0.07 \\ \hline
ResNet    & F1  & -0.32 & 0.14  & \textbf{0.69} & 0.12 & 0.19 & -0.16 & 0.09 & \textbf{0.85} & -0.07 & 0.31 & 0.03  & 0.15  & \textbf{0.70} & 0.01 & 0.29 \\
ResNet    & Pre & -0.30 & 0.12  & \textbf{-0.14} & -0.05 & -0.58 & -0.50 & 0.03 & \textbf{-0.16} & -0.11 & -0.18 & -0.58 & -0.12 & \textbf{0.00} & -0.14 & -0.05 \\
ResNet    & Re  & -0.16 & 0.11  & \textbf{0.70} & 0.12 & 0.36 & 0.05  & 0.10 &\textbf{0.79} & 0.01 & 0.27 & 0.24  & 0.20 & \textbf{0.60} & 0.08 & 0.21 \\ \hline
ViT16     & F1  & -0.52 & 0.25  & \textbf{0.56} & 0.14 & -0.04 & -0.51 & 0.20 & \textbf{0.65} & 0.04 & 0.34  & -0.36 & 0.23  & \textbf{0.61} & 0.09 & 0.38  \\
ViT16     & Pre & -0.12 & 0.24  & \textbf{-0.36} & -0.05 & -0.77 & -0.43 & 0.15 & \textbf{-0.45} & 0.01 & -0.16 & -0.56 & 0.01  & \textbf{-0.19} & -0.06 & -0.02 \\
ViT16     & Re  & -0.37 & -0.01 & \textbf{0.6} & 0.10 & 0.45 & -0.22 & 0.02 & \textbf{0.75} & 0.02 & 0.29 & -0.04 & 0.10  & \textbf{0.56} & 0.09 & 0.25 
\\ \hline \hline
\end{tabular}
}
\caption{Correlation Between Performance and Illumination Distribution Difference in CIFAKE1}
\label{tab:correlation_illumination}
\end{table*}

\subsection*{Differences in Model Learning Process}
\subsubsection*{Differences in Model Structure}
\begin{figure}
    \centering
    \includegraphics[width=1\linewidth]{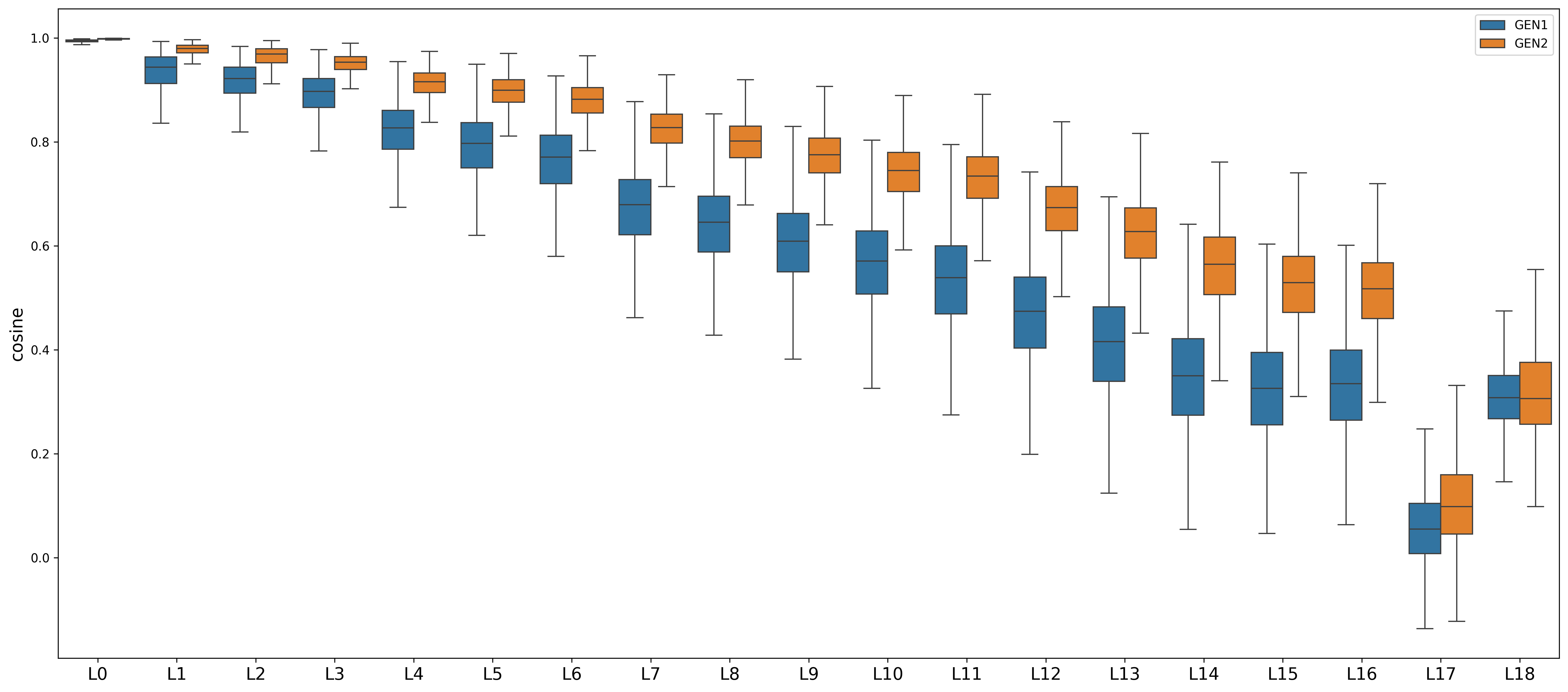}
    \caption{Comparison of cosine similarity based on output features from each layer of MobileNet, trained on REAL data: This shows the distribution of similarity across layers (L0–L18), visualizing feature variability and central tendencies.}
    \label{fig:comparison_cosine_similarity}
\end{figure}

The box plots of cosine similarity for each layer, shown in Figure \ref{fig:comparison_cosine_similarity}, demonstrated that CIFAKE1 already diverges from the REAL model starting from the feature layer immediately after the input layer. Significant differences are particularly evident in the early layers sensitive to low-level features such as color and shape, and this divergence persists through the 17th layer. Although similarity slightly increases in the final layer, it overall showed a declining trend, with wide variation. In contrast, CIFAKE2 shows no substantial change in cosine similarity relative to the REAL model until the early layers (approximately 3th layer). Even in the intermediate layers, the range of divergence in feature representations remains about half that of CIFAKE1. While divergence tends to increase as the model progresses to later layers, the rate of decline is less pronounced than in CIFAKE1. These trends were consistently observed across all models. 

\begin{figure}
    \centering
    \includegraphics[width=1\linewidth]{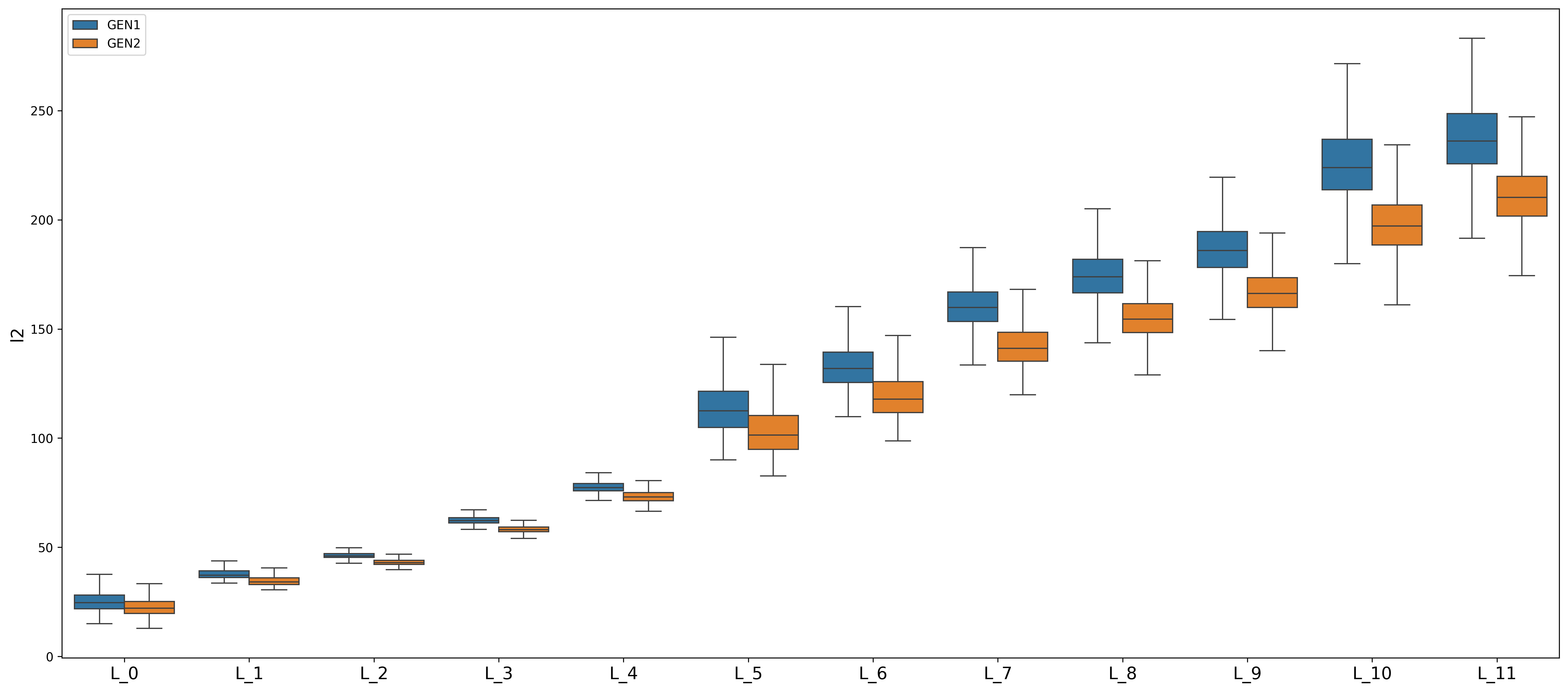}
    \caption{Comparison of L2 Distance based on output features from each encoder of ViT16, trained on REAL data: This shows the distribution of similarity across encoder layers (L0–L18), visualizing attention map variability and central tendencies.}
    \label{fig:comparison_l2}
\end{figure}

As shown in Figure \ref{fig:comparison_l2}, the L2 distance also increases as the layer depth increases, becoming farther compared to the REAL model. However, the gap between the CIFAKE2 model and the REAL model is not as large as that between the REAL model and CIFAKE1.

\subsubsection*{Realistic Scenarios}
\textbf{One class replacement:} Table \ref{tab:one_class_replacement} shows the change in performance metrics when only the frog class is replaced with the REAL dataset. For all models, Recall for the frog class improved significantly, while Precision decreased substantially. As a result, although the F1 score showed slight improvement for some models, no significant overall improvement was observed on average.

\begin{table}[]
\resizebox{\columnwidth}{!}{%
\begin{tabular}{l|lll|lll|lll}
\\ \hline \hline
\multirow{2}{*}{Class} & \multicolumn{3}{|l}{ResNet50} & \multicolumn{3}{|l}{MobileNetV2} & \multicolumn{3}{|l}{ViT16}  \\
                       & PRE  & RE  & F1    & PRE   & RE   & F1     & PRE & RE & F1    \\ \hline
airplane               & 0.07 & -0.08   & -0.06 & 0.07        & -0.06    & -0.04  & 0.02      & -0.08  & -0.06 \\
automobile             & 0.04 & -0.17   & -0.12 & 0.02        & -0.05    & -0.04  & 0.03      & -0.11  & -0.05 \\
bird                   & 0.21 & -0.33   & -0.06 & 0.18        & -0.35    & -0.02  & 0.25      & -0.26  & 0.03  \\
cat                    & 0.12 & -0.24   & -0.15 & 0.04        & -0.20    & -0.14  & 0.10      & -0.32  & -0.19 \\
deer                   & 0.09 & -0.20   & -0.18 & 0.08        & -0.14    & -0.13  & 0.06      & -0.29  & -0.25 \\
dog                    & 0.07 & -0.28   & -0.16 & 0.03        & -0.25    & -0.15  & 0.09      & -0.35  & -0.20 \\
frog                   & -0.52 & 0.92    & 0.21  & -0.42  & 0.91  & 0.20   & -0.69     & 0.66   & -0.06 \\
horse                  & 0.14  & -0.15   & -0.02 & 0.18 & -0.09    & 0.01   & 0.05 & -0.05  & 0.01  \\
ship                   & 0.18  & -0.12   & 0.06  & 0.16 & -0.15    & 0.01   & 0.08 & -0.01  & 0.04  \\
truck                  & 0.07  & -0.10   & -0.01 & 0.13 & -0.17    & -0.01  & 0.06 & -0.06  & 0.02  \\ \hline
Mean                   & \textbf{0.05} & \textbf{-0.08} & \textbf{-0.05} & \textbf{0.05} & \textbf{-0.05} & \textbf{-0.03}  & \textbf{0.01} & \textbf{-0.09}  & \textbf{-0.07} \\
\hline \hline
\end{tabular}
}
\caption{Improvements by One-class replacement}
\label{tab:one_class_replacement}
\end{table}

\begin{figure*}
    \centering
    \includegraphics[width=1\linewidth]{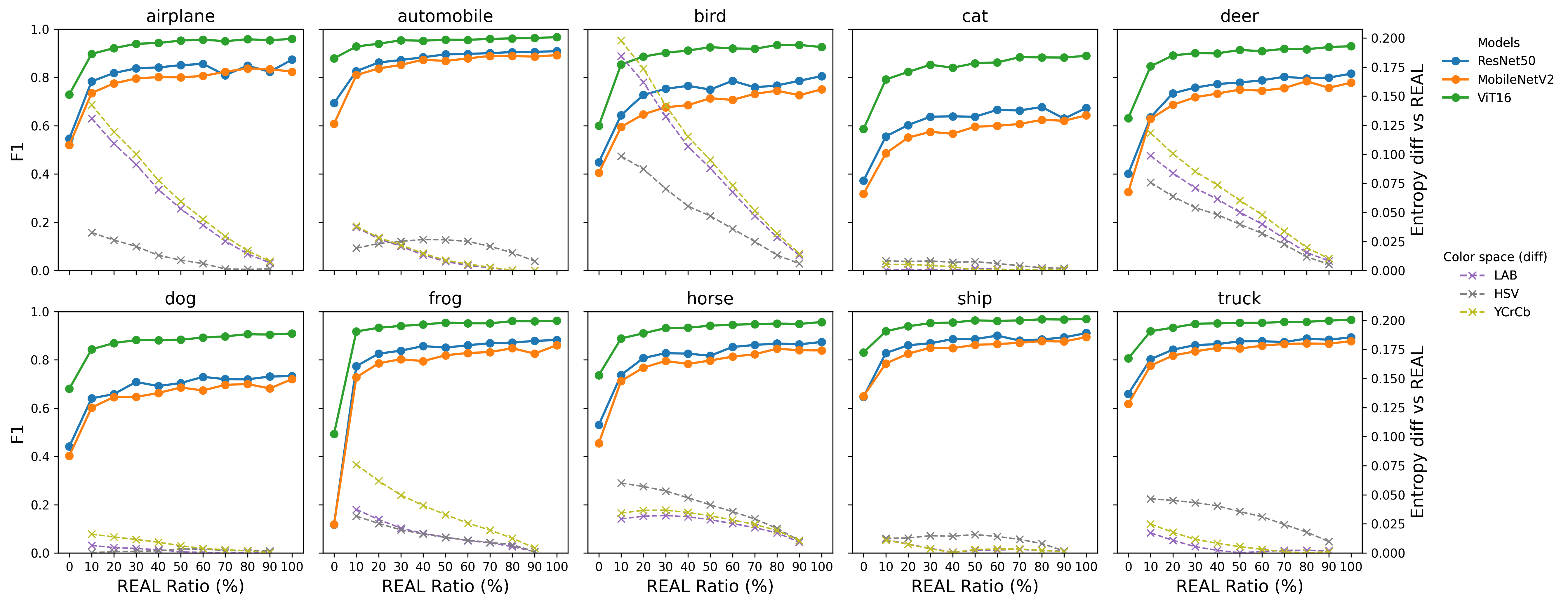}
    \caption{F1 score progression when mixing REAL images into the CIFAKE1 dataset: 0\% indicates CIFAKE1 only; subsequent values show REAL image proportions increasing in 10\% increments, with 100\% indicating REAL data only. The line graph of the cross markers show how the entropy statistics of the three types of illuminance channels transitioned.}
    \label{fig:f1score_progression}
\end{figure*}

\textbf{Randomize Mixture:} As shown in Figure \ref{fig:f1score_progression}, the F1 score for frogs, which was significantly low when using only CIFAKE1, improved substantially with just a 10\% mixture of REAL data. Furthermore, across all models, when the mixture ratio is around 30–40\%, no significant difference is observed compared to using REAL data alone.

It was confirmed that the statistical illuminance entropy value also approximates the REAL value as the number of REAL images increases. Similar to the performance improvement, even adding a small amount of REAL data significantly improved the statistical information.

\begin{table}[]
\begin{tabular}{l|llllll|llllll}
\hline \hline
& \multicolumn{6}{l|}{MobileNetV2} & \multicolumn{6}{l}{ViT16} \\
Label &  Ep 0   &   10  &  20   &  100   &  150  & Real & Ep 0   &   10  &  20   &  100   &  150  & Real  \\ \hline
airplane & 0.52 & 0.70 & 0.70 & 0.73 & 0.73 & \textbf{0.82} & 0.73 & 0.92 & 0.91 & 0.92 & 0.93 & \textbf{0.96}  \\
automobile & 0.61 & 0.82 & 0.81 & 0.83 & 0.82 & \textbf{0.89} & 0.88 & 0.93 & 0.94 & 0.94 & 0.94 & \textbf{0.97}  \\
bird & 0.41 & 0.54 & 0.58 & 0.58 & 0.57 & \textbf{0.75} & 0.60 & 0.85 & 0.87 & 0.87 & 0.88  & \textbf{0.93}  \\
cat & 0.32 & 0.49 & 0.52 & 0.52 & 0.53 & \textbf{0.64} & 0.59 & 0.79 & 0.78 & 0.80 & 0.82 & \textbf{0.89} \\ 
deer & 0.33 & 0.59 & 0.64 & 0.64 & 0.65 & \textbf{0.78} & 0.63 & 0.83 & 0.86 & 0.85 & 0.86 & \textbf{0.93} \\
dog & 0.40 & 0.59 & 0.62 & 0.61 & 0.59 & \textbf{0.72} & 0.68 & 0.84 & 0.85 & 0.85 & 0.86 & \textbf{0.91} \\ 
frog & 0.12 & 0.67 & 0.74 & 0.74 & 0.74 & \textbf{0.86} & 0.49 & 0.92 & 0.91 & 0.92 & 0.93 & \textbf{0.96} \\
horse &  0.46 & 0.73 & 0.72 & 0.73 & 0.73 & \textbf{0.84} & 0.74 & 0.88 & 0.90 & 0.89 & 0.91 & \textbf{0.96}  \\
ship& 0.65 & 0.78 & 0.79 & 0.81 & 0.80 & \textbf{0.89} & 0.83 & 0.94 & 0.94 & 0.94 & 0.94 & \textbf{0.97} \\
truck &  0.62 & 0.77 & 0.78 & 0.80 & 0.79 & \textbf{0.88} & 0.81 & 0.93 & 0.94 & 0.93 & 0.94 & \textbf{0.97} \\
\hline \hline
\end{tabular}
\caption{Results of additional fine-tuning of MobileNetV2 and ViT16 using 10\% of real data, evaluated from 10 to 150 epochs.}
\label{tab:additional_finetune}
\end{table}

\textbf{Additional Finetuning with Small Real Data:} Table \ref{tab:additional_finetune} shows the F1 score performance when training for an additional 150 epochs using 10\% REAL data. The score was evaluated every 10 epochs, and, as shown in Table \ref{tab:additional_finetune}, after a sharp increase at 10 epochs, it was only a slight increase. Even after training for up to 150 epochs, the results hardly improved and never exceeded those obtained from training on real data. This same result was observed for both CNN and transformer-based models.

\section*{Discussion}
In the discussion, we address each research question based on the experimental results from the previous section. \textbf{RQ1:} The difference between CIFAKE1 and CIFAKE2 in the high-dimensional feature space was confirmed to lie in the distribution of feature clusters and could be quantitatively captured as the distance between the clusters. It became apparent that a greater distance from the centroid within the REAL cluster negatively impacts classification performance. Comparing feature vector extraction using CLIP versus the DINO family, the DINO-based approach detected image-appropriate feature differences more clearly. While CLIP emphasizes semantic information through learning from text-image correspondence, DINO, which performs self-supervised learning using only images, captures visual features such as shape and texture more accurately. 

\textbf{RQ2:} The distribution differences between CIFAKE1 and 2 in low-level statistical information (in illuminance channels) were observed significantly in standard deviation, skewness, and entropy. The correlations between F1 scores and illuminance channel entropies in all color spaces were strong. These results indicate that classes with small differences in illuminance distribution compared to REAL images can be classified accurately, but as the differences increase, false negatives become more likely.

\textbf{RQ3:} Analysis of layers in the model training process also revealed that CIFAKE1 exhibited significant differences from REAL already during the initial layers in both cosine similarity and L2-distance, in other words, the stage of learning low-level statistical features. CIFAKE2 also gradually shows differences from REAL, and, while consistent with the argument that significant accuracy degradation occurs in the final layer \cite{hennicke2024mind}, depending on data quality, performance degradation can appear as early as the initial layers due to differences in color-space-level distributions, as seen in CIFAKE1. This indicates that the primary cause of performance degradation cannot be definitively attributed to the "final layer."

\textbf{RQ4:} Under realistic synthetic data scenarios, when images with differing brightness distributions, already low-level in CFAKE1, were used during initial fine-tuning, approximately 30\% to 40\% of REAL images could be secured. This achieved a distribution comparable to REAL training, with performance remaining nearly equivalent. Conversely, with only 10\% of the data, the distribution differences lingered, and even after approximately three times the fine-tuning, performance remained far below that of REAL. Furthermore, replacing only the poorly performing classes with REAL data disrupted the overall learning balance and did not improve performance.

These findings demonstrate that when applying synthetic data for fine-tuning, it is critically important to comprehensively measure the anticipated differences in properties compared to REAL data beforehand. Depending on these differences, training solely on synthetic data carries the risk of inducing data shifts. To mitigate these risks, a strategy of blending data to make the overall training dataset's characteristics approach those of REAL data could be adopted.

\subsection*{Limitations}
This study conducted experiments using publicly available general-purpose natural image datasets. Therefore, the findings from these results may not be directly applicable to specialized domains that heavily rely on pigment distribution or shape information, such as medical images. Particularly in application areas where diagnostic accuracy and safety are strongly demanded, further domain-specific verification is necessary.

\subsection*{Future Work}
Based on the findings from this research, we are planning to propose a method for stable learning in distributed learning environments prone to instability due to heterogeneous data distributions by monitoring the quality of learning data.

\section*{Conclusion}
To meaningfully utilize synthetic images in realistic scenarios, it is crucial to pre-assessment of the quality of the data. This study demonstrated that the degree of performance degradation can be predicted by quantifying differences in high-dimensional feature space as feature vector distances using DINOv3, and further employing entropy differences in low-level statistical information regarding illuminance. Furthermore, results from data blending experiments simulating real-world deployment confirmed that incorporating approximately 30\% REAL data per class during training mitigates data distribution bias and maintains classification performance comparable to training solely on real images. These findings provide practical guidelines for pre-evaluating synthetic data of unknown quality and safely and effectively incorporating it into image classification, thereby contributing to enhanced reliability and safety when utilizing synthetic data.

\section*{Acknowledgments}
The authors would like to thank the EPSRC National Edge AI Hub at the University of Hull for its support throughout this research. The authors also acknowledge COST Action European Materials Informatics Network (EuMINe CA22143) for supporting a Short-Term Scientific Mission, which contributed to the completion of this work.

\bibliography{main}

\end{document}